\pdfoutput=1
\def\year{2020}\relax
\documentclass[letterpaper]{article} 
\usepackage{aaai20}  
\usepackage{times}  
\usepackage{helvet} 
\usepackage{courier}  
\usepackage[hyphens]{url}  
\usepackage{graphicx} 
\usepackage{graphicx}  
\newcommand{\citet}[1]{\citeauthor{#1}~\shortcite{#1}}
\usepackage{multirow,booktabs}

\title{An Empirical Study of Content Understanding in\\ Conversational Question Answering}
\author{Ting-Rui Chiang\quad Hao-Tong Ye\quad Yun-Nung Chen\\ 
National Taiwan University, Taipei, Taiwan\\ 
\{r07922052, r08922065\}@csie.ntu.edu.tw\quad y.v.chen@ieee.org 
}
\begin{document}
\maketitle
\begin{abstract}
  With a lot of work about context-free question answering systems, there is an emerging trend of conversational question answering models in the natural language processing field.
  Thanks to the recently collected datasets, including QuAC and CoQA, there has been more work on conversational question answering, and recent work has achieved competitive performance on both datasets.
  However, to best of our knowledge, two important questions for conversational comprehension research have not been well studied:
  1) How well can the benchmark dataset reflect models' content understanding? 
  2) Do the models well utilize the conversation content when answering questions?
  To investigate these questions, we design different training settings, testing settings, as well as an attack to verify the models' capability of content understanding on QuAC and CoQA.
  The experimental results indicate some potential hazards in the benchmark datasets, QuAC and CoQA, for conversational comprehension research. 
  Our analysis also sheds light on both what models may learn and how datasets may bias the models.
  With deep investigation of the task, it is believed that this work can benefit the future progress of conversation comprehension. The source code is available at \url{https://github.com/MiuLab/CQA-Study}.
\end{abstract}

\section{Introduction}

Answering questions in the conversational manner has become an important task for machine reading comprehension.
There are two benchmark conversational question answering datasets, QuAC \cite{choi2018quac} and CoQA \cite{reddy2018coqa}.
Different from traditional machine reading comprehension \cite{rajpurkar2016squad,nguyen2016ms,rajpurkar2018know} whose questions are context-free, questions and answers in QuAC and CoQA are collected in a conversational manner.
Same as original machine reading comprehension, questions related to a given passage are asked.
However, questions may be also related to the given conversational history, and should be answered accordingly.
Such conversational setting is regarded as more practical because people tend to seek for information in a conversational way.
QuAC and CoQA also feature many linguistic phenomena unique to conversations, so they are believed to be important materials for research about conversational question answering.

This work focuses on investigating \textit{how well the performance of a model on these two benchmark datasets reflects its capability of comprehension}.
If higher performance on these datasets does not necessarily imply better conversation comprehension, then further investigation must be done when models claim their better understanding performance.
However, it has not been well investigated by any of the prior work~\cite{choi2018quac,huang2018flowqa,zhu2018sdnet,yeh2019flowdelta}.

In this paper, we further analyze \textit{whether the recent models achieving competitive performance rely on content comprehension}.
It is motivated by the fact that the position of the answer to the previous question is widely utilized in many of previous conversational question answering models, such as BiDAF-with-ctx~\cite{seo2016bidirectional} and FlowQA~\cite{huang2018flowqa}.
Those models leverage the datasets' property that answers or rationals can always be found as a span of the given passage.
Thus those models can access the informative content provided by the position information.
Models are expected to learn to understand the conversation based on its content.
However, it is not clear whether the models rely on the previous conversation content or merely the position information.


The goal of this paper is to investigate the above two questions and to guide the future research on conversational question answering.
We focus on recent open-sourced models, including  FlowQA~\cite{huang2018flowqa}, BERT~\cite{devlin2018bert}, and SDNet~\cite{zhu2018sdnet}. 
Similar to \citet{sankar-etal-2019-neural}, we design a set of experiments consisting of different training settings to address the first question and different testing settings to address the second question.
The results express some concerns about conversational QA models:
1) Higher performance on QuAC and CoQA does not necessarily imply better content comprehension.
2) Models trained on QuAC show the tendency of heavily relying on the previous answers' positions rather than their textual content.
By pointing out these issues, some potential hazards may be avoided in the future research. 
Furthermore, our presented experiments can serve as an analysis tool for conversational question answering models in the future.

\section{Related Work}

Machine reading comprehension has attracted lots of interests in recent years.
Many datasets \cite{rajpurkar2016squad,rajpurkar2018know,nguyen2016ms,trischler2017newsqa,lai2017race,shao2018drcd} have been created, and there has been many models \cite{wang2016machine,weissenborn2017making,hu2018reinforced,xiong2018dcn,shen2017reasonet,liu2018stochastic} crafted for advancing the tasks.
The above datasets and models are for context-free question answering. 
Recently, with the trend of conversational interactions, people started to focus on conversational understanding. Therefore, datasets for conversational question answering were built for promoting this research direction \cite{saha2018complex,choi2018quac,reddy2018coqa}, and a lot of models \cite{huang2018flowqa,zhu2018sdnet} were proposed to tackle this challenge. 

Most investigation on what machine comprehension models learn are for those context-free question answering systems. \citet{jia2017adversarial} proposed a method to generate adversarial examples in order to test the model robustness. \citet{kaushik2018much} conducted experiments to verify the reading required to answer questions in the dataset. \citet{weissenborn2017making} found a feature indicating if a word appears in the question important, and suggested that questions can be answered with some rules that rely only on superficial features. 
\citet{rondeau2018systematic} validated the suggestion by a series of systematic experiments. 
However, the conversational question answering systems have been rarely explored.
Although \citet{yatskar2018qualitative} compared CoQA, SQuAD 2.0 and QuAC qualitatively, there was no investigation on what conversational question answering models capture.


The phenomenon that models does not utilize all useful features is common in diverse areas.
For example, \citet{nie2018analyzing} showed that natural language inference models with high accuracy relied much on the lexical-level features but utilized little compositional semantics.
\cite{sankar-etal-2019-neural} found that not whole conversation history is used in neural dialogue generation systems.
Similar phenomena happened in the computer vision area, where \citet{geirhos2018imagenettrained} indicated that CNN trained on ImageNet relied much on textual information rather than shape information, and \citet{brendel2018approximating} further showed that only textual information can achieve very high accuracy on ImageNet.
Our work extends a similar idea to conversational question answering systems and demonstrates that positional information rather than semantic information can be exploited to master some tasks.

\section{Conversational Question Answering}

\begin{figure}[t!]
    \centering
    \includegraphics[width=\linewidth]{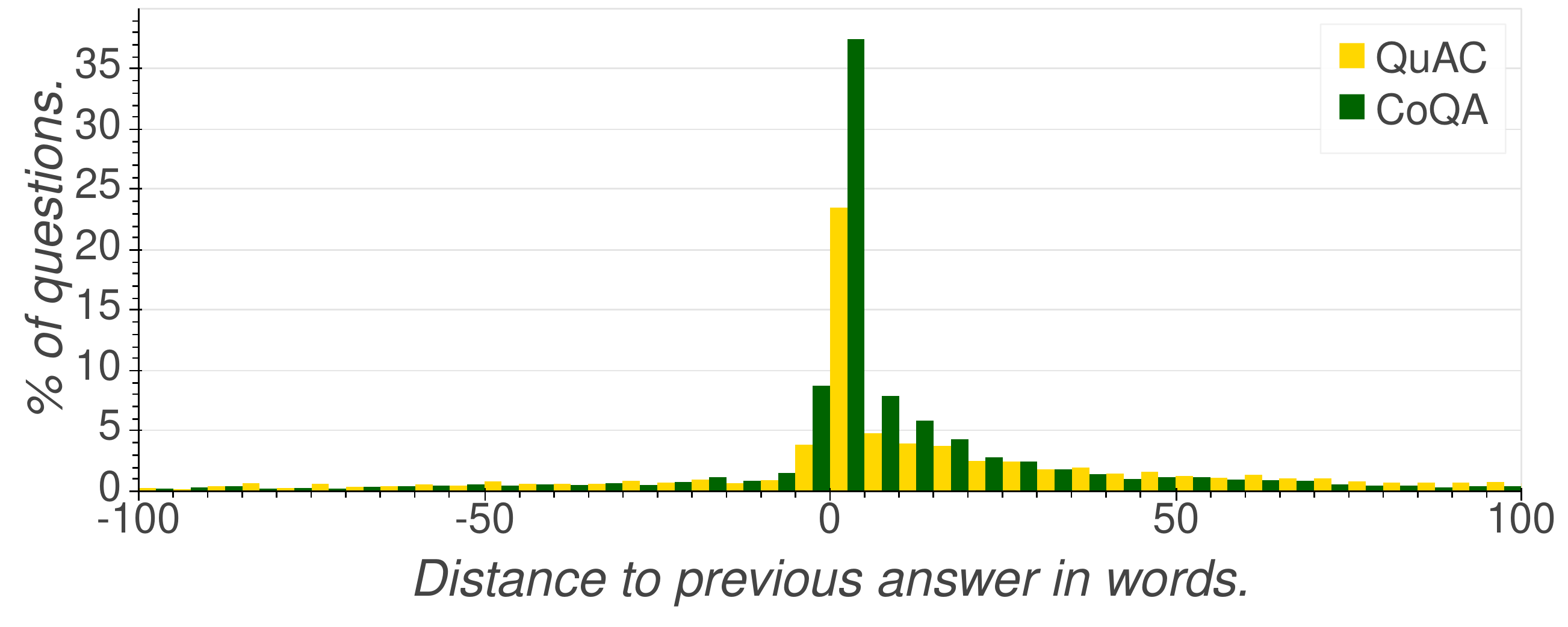}
    \caption{The histogram of distance between answers to consecutive questions of a conversation in words. The bin size is 5 words. The distance is counted as zero if the two answer spans are overlapped, and is positive if the current is after the previous answer, negative otherwise.}
    \label{fig:answer-distance}
\end{figure}

There are two main datasets for conversational question answering, QuAC~\cite{choi2018quac} and CoQA~\cite{reddy2018coqa}.
Questions and answers are collected in a conversational manner, where each conversation includes two participants: a student who asks question about a given passage, and a teacher who answers the question according to the passage.
The teacher in both sets may reply ``\textit{no answer}'' if the answer cannot be found in the given passage. 
When evaluating the model for both tasks, the content as well as the position of answers to the previous question is available to the model.
In the statistics of two datasets, answers to the consecutive questions tend to be close to each other depicted in Figure \ref{fig:answer-distance}.

\begin{figure}[t!]
    \centering
    \includegraphics[width=\linewidth]{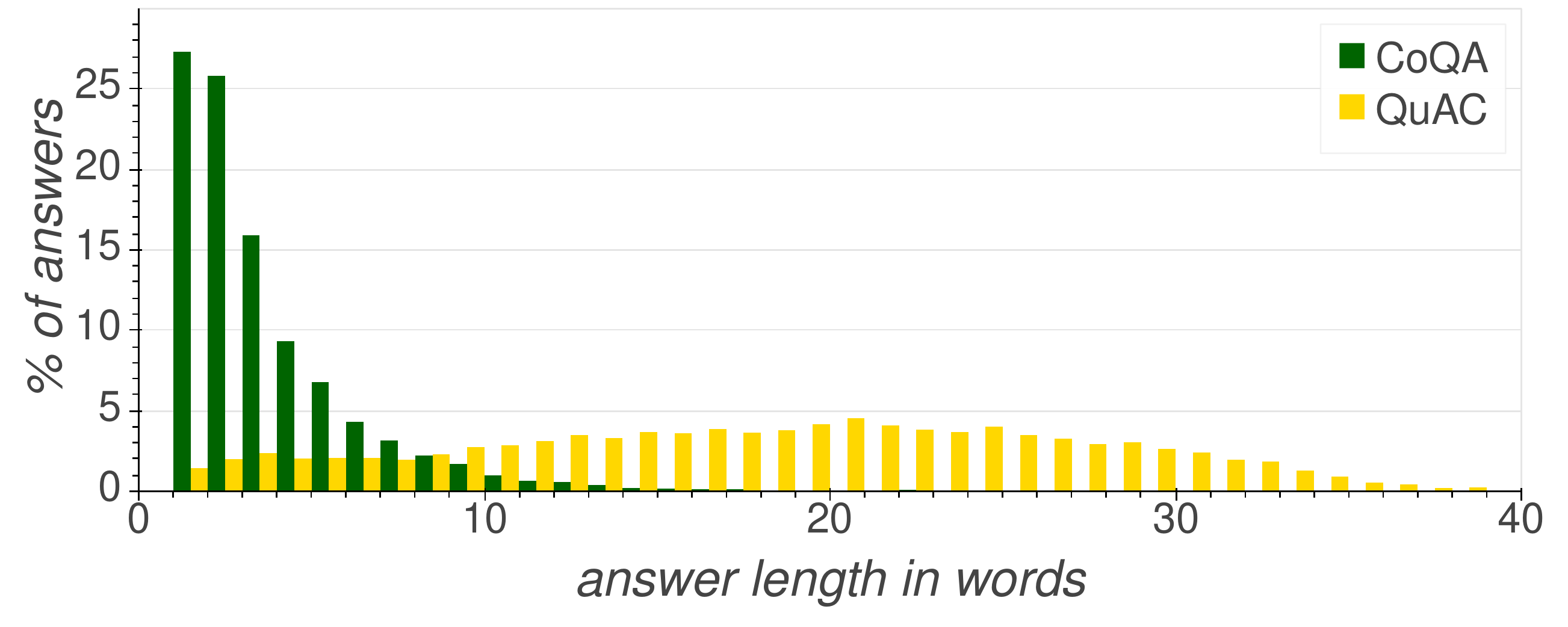}
    \caption{Histogram of answer length distribution in QuAC and CoQA. The bin size is 1 word.}
    \label{fig:answer-length}
\end{figure}

Even though both datasets are collected for conversational question answering, they have several different properties.
\begin{itemize}
    \item Answer format\\
    Answers in QuAC are always the text spans in the given passage, while answers in CoQA are free texts similar to some spans in the passage.
Answers in QuAC is generally longer than answers in CoQA shown in Figure~\ref{fig:answer-length}, where the distribution implies that QuAC is more realistic than CoQA.
Answering ``\textit{yes}'' or ``\textit{no}'' is also allowed in CoQA.
Note that the evidence span (span in the passage that supports the answer) is provided in CoQA, so the previous answers' position information is still available.
\item Dataset collection process\\
The Amazon mechanical turkers who generated the CoQA dataset have full access to the passage.
On the other hand, turkers who generated questions in QuAC cannot see the passages. 
The latter setting may be more suitable for practical applications, because real users want to seek for information using questions when not reading passages. 
\end{itemize}

\section{Models}
\label{sec:models}
We consider models including FlowQA, BERT, and SDNet, as they are the only publicly available models till now. 
For FlowQA and SDNet, we use the code released by the authors. Some modifications are made for the following experiments\footnote{For reproducibility, we will release all code, script, and experiment settings.}.
Each models of each setting are trained with 3 different random seeds, and the resulted mean and standard deviation value are reported for reliability.

All models in the experiments are mainly based on those designed for single-turn reading comprehension tasks, so this section focuses on describing the modification for each method in order to handle understanding in conversational question answering.
Below three models are detailed.

\subsection{FlowQA}
\label{sec:models-flowqa}
FlowQA~\cite{huang2018flowqa} is the model specifically designed for conversational question answering, which contains a mechanism that can incorporate intermediate representations generated during the process of answering previous questions.
This model significantly improved conversational machine comprehension tasks for both QuAC and CoQA data.
In FlowQA, the main mechanism designed for the conversational structure is the \textsc{Integration-Flow} layer. 
In the model, there is a question-aware context representation $C_i$ for each question $Q_i$ in the history, and the \textsc{Integration} process simply applies BiLSTM to each $C_i$ independently.
After that, the \textsc{Flow} process applies BiLSTM to the same word across different context representations in order to capture knowledge of previous questions.
Also, an \emph{one-bit} feature is added to the input context word representation indicating whether the word appears in previous answers. 
The rest of the reasoning procedure is almost identical to FusionNet \cite{huang2017fusionnet} that focuses on single-turn machine comprehension.

\subsection{BERT}
\label{sec:models-bert}
In the current state-of-the-art question answering models, most models leverage the benefits from BERT~\cite{devlin2018bert} to advance the task.
We apply BERT on QuAC by converting the task into a single-turn machine reading comprehension task such as SQuAD~\cite{rajpurkar2016squad}.
We prepend previous $N$ questions to the current question $Q_k$, so it becomes $\hat{Q_k} = \{Q_{k-N}, \dots, Q_{k-1}, Q_k\}$.
At the embedding layer, beside the original word embedding, segment embedding and position embedding, an additional embedding is also added to represent whether the word appears in previous answer spans. 
Then we follow the procedure of applying BERT to SQuAD that concatenates the extended and the context to form the input and uses the context output representations from BERT to predict the start and the end of the answer span \cite{devlin2018bert}.

\subsection{SDNet}
\label{sec:models-sdnet}
To incorporate the information from dialogue history, SDNet prepends not only previous questions but also previous answers to the current question $Q_k$, i.e., $\hat{Q_k} = \{Q_{k-N}, A_{k-N}, \dots, Q_{k-1}, A_{k-1}, Q_k\}$, and no more additional effort is put to deal with the conversational structure.
For the input representation for both context and question words, they used BERT as contextualized embeddings along with GloVe. The rest of the model architecture for reasoning is highly inspired by FusionNet~\cite{huang2017fusionnet}.

\section{How Well the Performance Reflects Content Comprehension?}
\label{sec:necessity}

This section attempts at investigating how well the performance reflects the capability of comprehension on QuAC and CoQA?
Both datasets claim rich linguistic phenomena unique to conversations, where QuAC claims to have 61\% of questions including coreference referring to entities in the given passage, 44\% of coreference referring to entities in previous history, and 11\% of questions that ask for more information in the conversation. 
Also, CoQA claims to have 49.7\% of questions with explicit coreference to conversations, and 19.8\% with implicit ones. 
Given such high ratio of questions related to conversations, it is natural to expect that \emph{higher performance implies better understanding in conversational question answering}.
Especially, the understanding should be based on the content of the conversation, in which those special linguistic phenomena is embodied.
To inspect the expectation, we design experiments based on the premise: 
If comprehension on the content of conversation is reflected well by the performance, then model trained without the access to conversation content should not achieve high performance.

\subsection{Experimental Settings}
\label{sec:necessity-exp}
We compare models trained and tested with three different settings:
\begin{itemize}
    \item Original: The model has free access to the previous conversation history, as the setting proposed by the models.
    \item \textsf{- text}: The model has no access to the content of the answer to the previous question, but has access to their position in the provided context. The previous questions are not used either.
    \item \textsf{- conversation}: The model has \textit{no any} access to the previous conversation history.
\end{itemize}

In the \textsf{- text} setting, the answer span in the passage is masked with zeros. 
As questions are to seek for information, the answer content should be highly informative. 
Therefore, the information loss by answer masking cannot be easily compensated by the surrounding words.
Especially, for QuAC, since answer spans are typically as long as sentences, masked language models like BERT can \textit{in no way} recover the masked part.
Thus, in this setting, the only information about previous answers is their positions in the passage. 

\begin{table}[t!]
\centering

\begin{tabular}{l |l |l}
\toprule
Dataset               & Model                    & F1 (stdev) \\
\midrule
\multirow{6}{*}{QuAC} & FlowQA                   & 64.4 (.30) \\
                      & BERT                     & 63.6 (.16) \\
                      \cmidrule{2-3}
                      & FlowQA - text            & 62.4 (.20) \\
                      & BERT - text              & 62.2 (.48) \\
                      \cmidrule{2-3}
                      & FlowQA - conversation    & 54.1 (.13) \\
                      & BERT - conversation      & 55.3 (.03) \\
\midrule
\multirow{6}{*}{CoQA} & FlowQA                   & 76.9 (.22) \\
                      & SDNet + position         & 76.4 (.31) \\
                      \cmidrule{2-3}
                      & FlowQA - text            & 71.5 (.24) \\
                      & SDNet + position - text  & 74.0 (.19) \\
                      \cmidrule{2-3}
                      & FlowQA - conversation    & 63.4 (.11) \\
                      & SDNet - conversation     & 68.3 (.45) \\
\bottomrule
\end{tabular}
\caption{Model performance on the validation set of QuAC and CoQA. Note that the original SDNet model does not utilize the position information, so \textit{SDNet + position - text} is a modified SDNet with additional one dimension feature indicating the previous answer as the input.}
\label{tab:necessity-exp-result}
\end{table}

\subsection{Discussion}
We compare the results of models in Table~\ref{tab:necessity-exp-result}.
For both QuAC and CoQA, models trained without access to conversation \textit{content} can achieve performance significantly better than models trained without access to any conversation history.
Especially for QuAC, \textsf{- text} models consistently outperform \textsf{- conversation} models by up to absolute 7\% F1 score.
As for CoQA, though not as consistently, but similar comparison can also be observed.

The above results indicate that better understanding in the dataset is not well reflected by the performance on these two datasets.
Undoubtedly, better comprehension should be based on semantic understanding specific to the textual content.
However, even no content is provided to \textsf{- text} models, \textsf{- text} models can still achieve performance higher than \textsf{- conversation} models.
It indicates that \textit{higher performance} does not necessarily imply \textit{better content understanding}.
Therefore, future models may need to further verify whether they indeed focus on semantic understanding instead of utilizing the position information only.

\section{Do Models Understand Conversation Content?}

The results in \S\ref{sec:necessity} do not answer the question that if the full models understand content of conversations well.
As shown in \S\ref{sec:necessity}, the full models can achieve better performance than models use only the answers' position information.
However, \S\ref{sec:necessity} also shows the usefulness of the previous position information.
Since the full model has access to both the position and content of the previous answers, it is not clear whether the better performance is contributed by understanding conversation content.
To answer this question, we analyze the trained models by a series of testing settings.

\subsection{Repeat Attack}
\label{sec:repeat-attack}

\begin{figure}[t!]
    \centering
    \includegraphics[width=\linewidth]{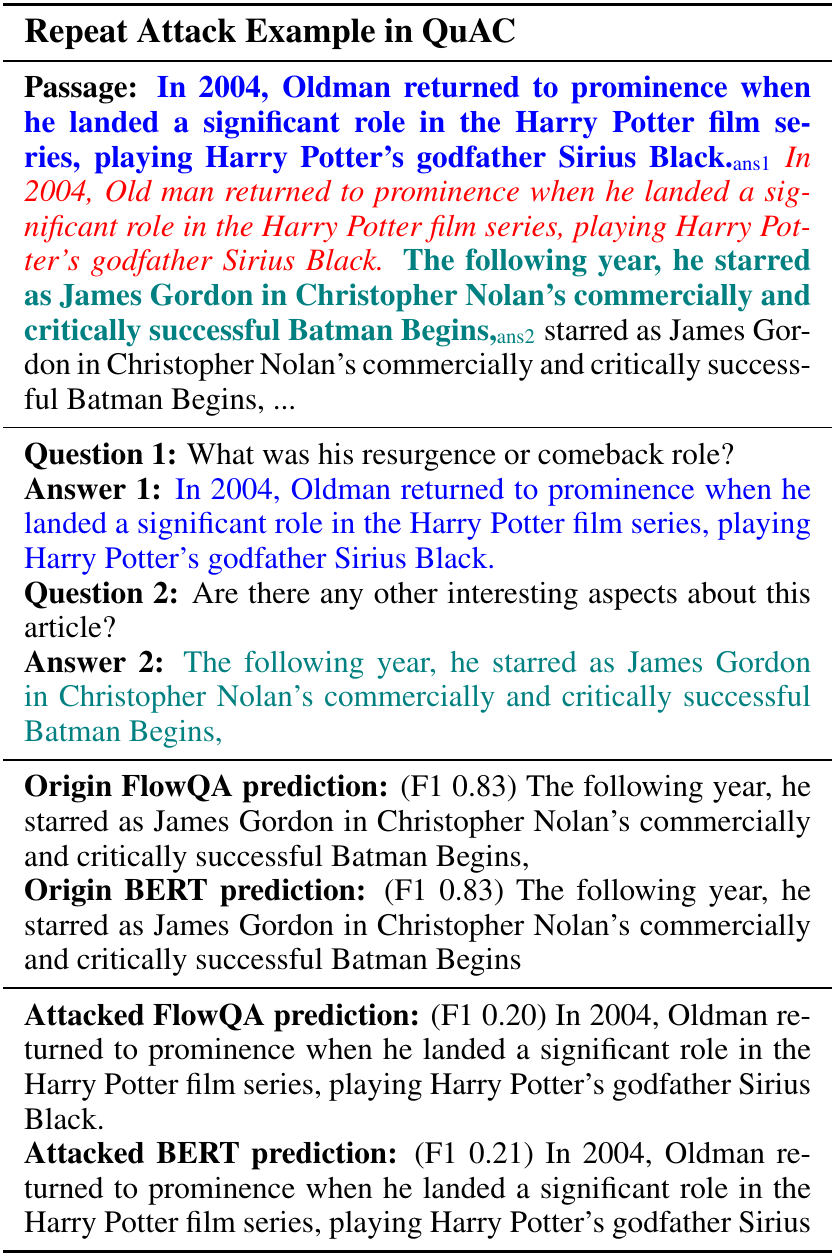}
    \caption{An example of repeat attack on QuAC and the model results. The red italic text is the inserted attack.}
    \label{tab:attack-quac}
\end{figure}

\begin{figure}[t!]
    \centering
    \includegraphics[width=\linewidth]{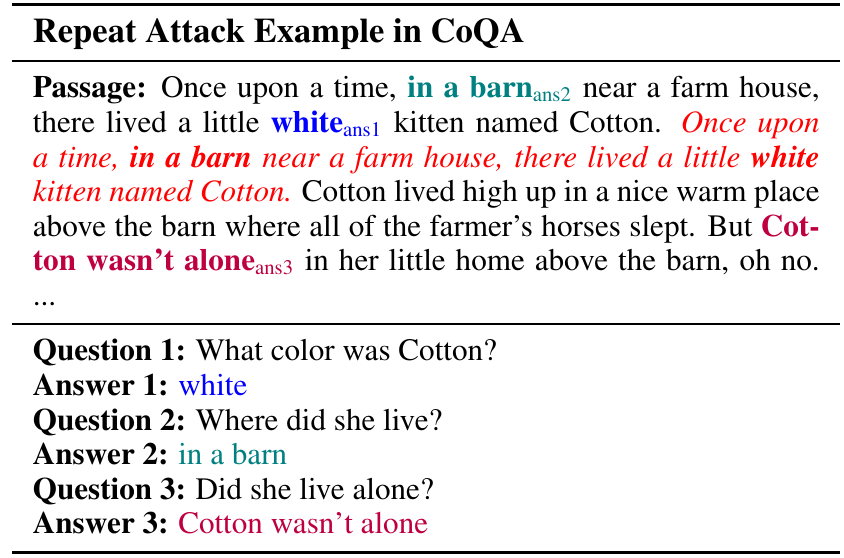}
    \caption{An example of repeat attack on CoQA. The red italic text is the inserted attack. In this example, the attack increases the distance between answer 2 and 3.}
    \label{tab:attack-coqa}
\end{figure}

\begin{figure}[t!]
    \centering
    \includegraphics[width=\linewidth]{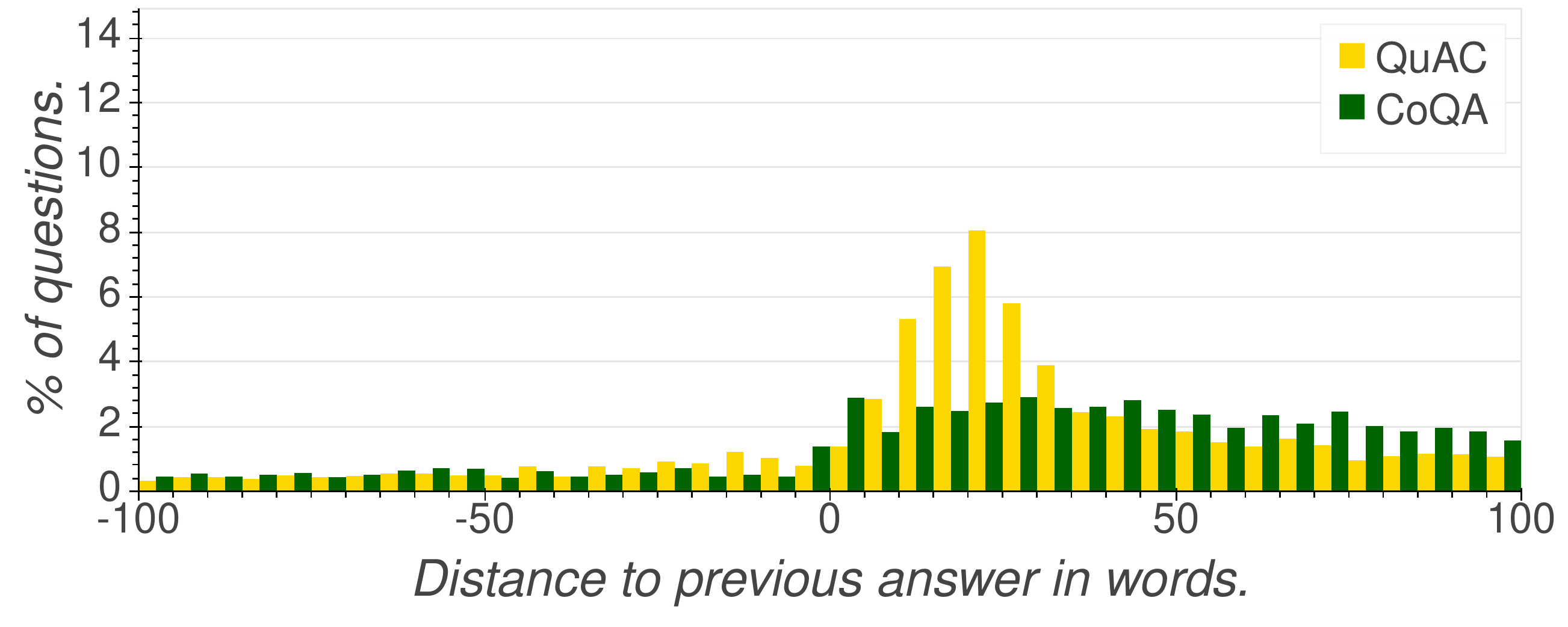}
    \caption{Histogram of distance between answers to consecutive questions in words \emph{after} the attack. The bin size is 5. }
    \label{fig:answer-distance-with-attack}
\end{figure}

We propose \textit{repeat attack} that increases the distance between answers in the context.
To do so, a text span is repeated between the answer spans in the passage.
For QuAC, as most answers are sentences, we repeat the answer sentence for each answer spans in the passage.
For CoQA, since its answer span is generally much shorter than QuAC, we repeat sentences that contain the answer span.
The attack examples for both datasets are shown in Table~\ref{tab:attack-quac} and Table \ref{tab:attack-coqa}.
By repeating part of the passage, the meaning delivered should remain the same. 
When evaluating the models, the previous answer positions provided to models contain only the text as in original answer.
Due to the repeated text, the distance between consecutive answer spans is lengthen by this attack.
The distribution of the distance is visualized in Figure~\ref{fig:answer-distance-with-attack}, which is much smoother than the one before the attack (Figure~\ref{fig:answer-distance}).

We use \textit{repeat attack} to investigate models' understanding of conversation content.
It is motivated by the high ratio of answers close to answers to the previous questions (as shown in Figure~\ref{fig:answer-distance}).
It is possible that positions of previous answers leak the position information of the current answer.
The model may thus learn to take as the answer candidates the sentences close to the previous answer span.
On the other hand, if a model does answer the question by understanding answer content, then the model should be robust to this attack.
Therefore, by using the attacked data to test models that are trained on the normal data, the models' capability of understanding can be well investigated.

\begin{figure*}[t!]
    \centering
    \includegraphics[width=0.49\linewidth]{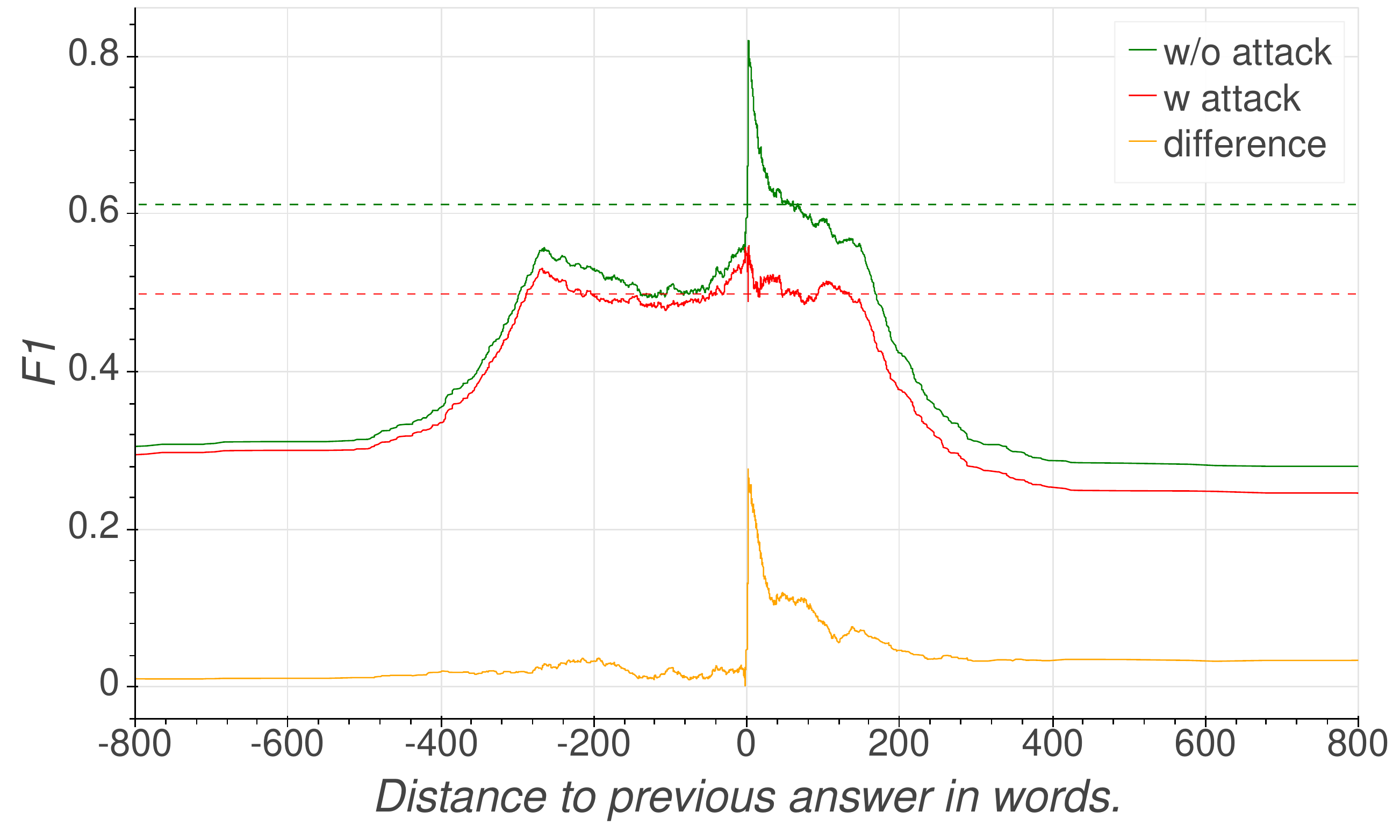}
    \includegraphics[width=0.49\linewidth]{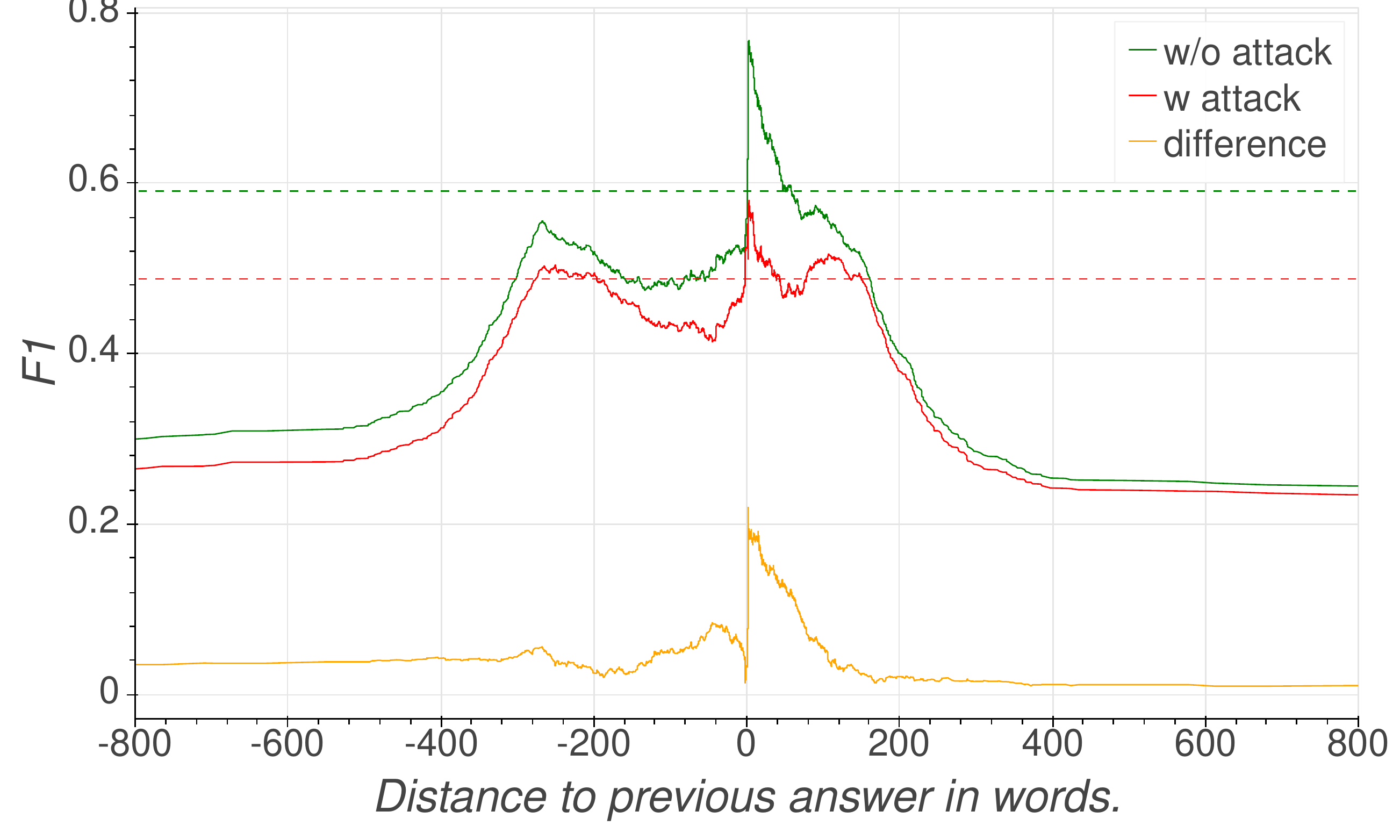}
    \caption{F1 change between before and after repeat attack in terms of the answer's distance to its previous answers on QuAC (left: FlowQA; right: BERT). The x-coordinate is the answer's distance to the previous answer in words, and the meaning of zero, positive and negative is same as in Figure \ref{fig:answer-distance}. The y-coordinate is the average F1 score. ``\textit{no answer}'' sample are ignored here. The dotted line is the average F1 score of all samples.}
    \label{fig:attack-curve}
\end{figure*}

For QuAC, the results shown in Table \ref{tab:repeat-exp-result} indicates that both FlowQA and BERT are sensitive to the distance between consecutive answers. 
Their performance drops significantly when applying the repeat attack.
The F1 score under attack is roughly the F1 score of models trained without any conversational information. 
Furthermore, we conduct the same experiments on FlowQA and BERT trained without using position information of the previous answers, and find that although they perform much worse than the full models, they are more robust against the attack.

To better investigate how the answer position is related to the model robustness, we plot the relation between F1 scores before/after attack in terms of the distance to previous answer in Figure \ref{fig:attack-curve}. 
We find that both BERT and FlowQA predict answers more accurately when the current answer has short distance to the previous answer (green and red lines). 
Also, our attack is more effective when the distance is short (yellow line). 
These findings imply that models trained on QuAC with the position information may rely less on the answer content but rely too much on the answer position.
In addition, models trained without the position information may rely more on the semantic information in the conversations.

We also investigate the performance affected by the attack based on if the question is a followup of the previous question, which is annotated in QuAC dataset. The followup questions are questions more likely to depend on the previous question, and therefore are expected to require understanding of conversation content. However, results shown in table \ref{tab:attack-followup} indicate that they are more vulnerable to our attack.

QuAC models' reliance on position information can be further shown by the qualitative error analysis on the attacked validation set. 
We inspect the questions that the models originally predict the answer with a high F1 score, but predict the answer with a low F1 scores when applying attack.
One sample is shown in Table~\ref{tab:attack-quac}, where two models under attack directly predict the sentence after the previous answer span regardless of the content.
On other cases, we find FlowQA under attack makes mistakes by selecting the next span more consistently.
In contrast, BERT under attack is prone to predict a much implausible answer or reply ``\textit{no answer}''. The reason may be that BERT is specialized to search answers in the region near the previous answer, so consequently, when the next sentence after the previous answer is incorrect, it will fail to answer the question correctly.

On the other hand, it is less clear if the drop of performances on CoQA dataset is due to insufficient of conversation content understanding.
Unlike on QuAC, \textsf{- position} models are not more robust than the original models as obviously as on QuAC.
By plotting figures for CoQA in the same setting, we find that questions in CoQA seem to be susceptible to the attack equally regardless of the distance to previous answers.
Conversation content understanding of the models trained on CoQA may require further investigation.

\begin{table}[t!]
\centering
\begin{tabular}{c |l | c | c }
\toprule
\multirow{2}{*}{Dataset} & \multirow{2}{*}{Model} & \multicolumn{2}{c}{Repeat Attack}  \\
                         &                        & w/o & w/ \\
\midrule
\multirow{4}{*}{QuAC} & FlowQA            & 64.4 (0.30) & 53.3 (0.75) \\
                      & BERT              & 63.6 (0.16) & 55.3 (0.74) \\
                      & FlowQA - position & 59.3 (0.37) & 56.9 (0.37) \\
                      & BERT - position   & 58.0 (0.18) & 56.5 (0.22) \\
\midrule
\multirow{4}{*}{CoQA} & FlowQA            & 76.9 (0.22) & 72.0 (0.20) \\
                      & SDNet             & 76.4 (0.31) & 71.8 (0.48) \\
                      & FlowQA - position & 76.8 (0.42) & 72.4 (0.24) \\
                      & SDNet - position  & 75.7 (0.76) & 70.9 (0.10) \\
\bottomrule
\end{tabular}
\caption{F1 score on the validation of QuAC and CoQA with or without attack.  "\textit{- position}" indicates training without the position information of answers to previous questions.}
\label{tab:repeat-exp-result}
\end{table}

\begin{table}
\centering
\begin{tabular}{l c | c | c}
\toprule
Model &       & Followup & No Followup   \\
\midrule
\multirow{3}{*}{FlowQA} & no attack & 64.0 (0.59) & 64.8 (0.11) \\
                        & attack    & 49.5 (0.59) & 57.6 (0.94) \\
                        & $\Delta$  & 14.6 (1.17) & 7.2 (1.00) \\                        
\midrule
\multirow{3}{*}{BERT}   & no attack & 63.4 (0.32) & 64.0 (0.57)   \\
                        & attack    & 52.4 (1.37) & 58.5 (0.18)   \\
                        & $\Delta$  & 11.0 (1.12) & 5.5 (0.41) \\
\bottomrule
\end{tabular}
\caption{Impact on F1 score by the attack.}
\label{tab:attack-followup}
\end{table}

\subsection{Predict without Previous Answer Text}
\label{sec:predict-with-mask}

To investigate if the content of previous answers is used by the models trained with position information, we measure the performance on the validation set predicted without the content of previous answers. 
To remove the content information of previous answers, we mask the previous answer spans with zeros.
Different to previous \textsf{- text} settings, models here has access to previous questions.
Particularly, for FlowQA, the content information of the previous answer may be flowed along with the flow structure, and the RNN memory of the previous answer spans is also reset to zeros.

The more performance drops when masking previous answers implies that the model relies more on the content information of those answers.
According to the results in Table \ref{tab:mask-exp-result}, it seems that all models more or less rely on the text of previous answers. 
Meanwhile, as expected, the models except \textsf{SDNet - position} trained without the position information almost drop to the performance of ones trained without any conversation history information.
\textsf{SDNet - position} is an exception because answers to previous questions are prepended to the question when training and testing, so masking answers in the passage does not remove all the content of previous answers.
However, it is surprising that FlowQA on QuAC can still keep the performance up to 60\% F1, implying that FlowQA may rely on position information much more than the semantic information.

\begin{table}[t!]
\centering
\begin{tabular}{l |l | c | c }
\toprule
\multirow{2}{*}{Dataset} & \multirow{2}{*}{Model} & \multicolumn{2}{c}{Ans. Mask}  \\
                         &                        & w/o & w/ \\
\midrule
\multirow{4}{*}{QuAC} & FlowQA            & 64.4 (0.30) & 60.5 (0.54) \\
                      & BERT              & 63.6 (0.16) & 52.6 (1.76) \\
                      & FlowQA - position & 59.3 (0.37) & 55.0 (1.80) \\
                      & BERT - position   & 58.0 (0.18) & 50.1 (0.45) \\
\hline
\multirow{4}{*}{CoQA} & FlowQA            & 76.9 (0.22) & 71.2 (0.41) \\
                      & SDNet             & 76.4 (0.31) & 73.5 (0.51) \\
                      & FlowQA - position & 76.8 (0.42) & 68.2 (0.28) \\
                      & SDNet - position  & 75.7 (0.76) & 75.6 (0.69) \\
\bottomrule
\end{tabular}
\caption{F1 results on the validation sets of QuAC and CoQA. Models are infered without/with applying masks on the previous answers. Models without suffix are trained with full access to conversation, while ``\textit{- position}'' indicates that the models are trained without the previous answer position information. Note that in both training settings, masks are not applied.}
\label{tab:mask-exp-result}
\end{table}

\subsection{Predict without Previous Answer Position}
\label{sec:predict-without-position}

To directly test to what extent the models rely on position information of previous answers, we conduct the experiments of predicting answers without position information.
The results are shown in Table~\ref{tab:position-exp-result}, where both results of FlowQA and BERT on QuAC drop significantly if not using position information.
Among them, FlowQA drops even more, regardless of the flow structure that models the dialog flow.
The performance is even lower than the models trained without using any conversation history.
It indicates that although the models on QuAC may rely on the semantics of previous answers, the position information is indeed exploited.
On the other hand, it is interesting to see that FlowQA trained on CoQA rely little position information of the previous answers.

\subsection{Implication of Above Experiments}

Our results show that the models trained on QuAC have high tendency to rely heavily on the position of previous answers. The proposed attack and experiment settings can serve as a diagnosis tool in the future.

\begin{table}[t!]
\centering
\begin{tabular}{l |l | c | c }
\toprule
\multirow{2}{*}{Dataset} & \multirow{2}{*}{Model} & \multicolumn{2}{c}{Position Info.}  \\
                         &                        & w/ & w/o \\
\midrule
\multirow{2}{*}{QuAC} & FlowQA            & 64.4 (0.30) & 48.1 (0.72) \\
                      & BERT              & 63.6 (0.16) & 54.9 (0.08) \\
\hline
\multirow{2}{*}{CoQA} & FlowQA            & 76.9 (0.22) & 76.7 (0.36) \\
                      & SDNet             & 76.4 (0.31) & 73.7 (0.18) \\
\bottomrule
\end{tabular}
\caption{F1 score on the validation sets of QuAC and CoQA. Models are inferred with/without access to position information, but trained with access to position information.}
\label{tab:position-exp-result}
\end{table}

\section{Dataset and Model Analysis}
\label{sec:why-coqa}

To further investigate the difference between models trained on QuAC and CoQA, two questions are focused here.

\paragraph{Why do CoQA models rely less on position information?}

It is unclear why the models trained on CoQA rely less on the position information of the previous answer as shown in the previous sections.
Figure \ref{fig:answer-length} shows that answers in CoQA are much shorter than in QuAC, so if we normalize the distance to the previous answer related to the length of the answers, the average distance to the previous answer in CoQA would be much longer than QuAC.
Furthermore, short answers also imply that an answer can be identified as the sentence containing the answer. 
To verify the above claim, we randomly shuffle the sentences in the passage for each CoQA example, so that the cross-sentence information and the sentence order are removed from the passage.
Then we use CoQA models trained without position information for answer prediction. 
In Table \ref{tab:result-coqa-shuffle}, roughly speaking, up to 70\% questions can still be answer correctly.
It thus supports our claim and partly explains why models trained on CoQA rely less on the position information.

\begin{table}[t!]
    \centering
    \begin{tabular}{c | l|cc}
    \toprule
    \multirow{2}{*}{Dataset} & \multirow{2}{*}{Model} & \multicolumn{2}{c}{Shuffle} \\
                             & & w/o & w/\\
    \midrule
    \multirow{2}{*}{CoQA} & FlowQA - position & 76.8 (0.42) & 71.7 (0.25) \\
                          & SDNet - position  & 75.7 (0.76) & 70.1 (0.59) \\
    \bottomrule
    \end{tabular}
    \caption{F1 score of models on shuffled validation set of CoQA. Models here do not use the position information of the previous answers.}
    \label{tab:result-coqa-shuffle}
\end{table}

\paragraph{Why do QuAC models rely more position information?}
We describe the potentail reason why the models trained on QuAC rely much on the position information. According to the analysis on QuAC \cite{choi2018quac}, 11\% of the questions is of the type ``\textit{Anything else?}''.
It is common in the general article where the information of the same type is written in near contexts.
Because the question ``\textit{Anything else?}'' is asked to seek for more information similar to the previous answer, it is very likely that the position of the previous answer provides a strong hint for the current answer. 
Though there is a high percentage of the questions containing pronouns in QuAC, they do not necessarily force the model to learn coreference resolution either. 
For the pronouns in questions, they often refer to entities in the previous answer. 
Because entities in consecutive sentences seldom change much, simply looking for the answer near the previous answer location may be sufficient to answer the question. 
Especially, we find personal pronouns in QuAC questions of a conversation often refer to only one same person.
Also, the referred person is often the main role in the passage.
This further removes the necessity to understand the conversation history.

\section{Conclusion}

This paper investigates content understanding of different models learned from different datasets. 
The experiments shows concerns:
1) Performance on QuAC and CoQA does not well reflect the model's comprehension on conversation content. 2) The model trained on QuAC does not necessarily learn conversation comprehension. 3) In CoQA, cross-sentence information is not that important for the current model.
By pointing out these concerns, we suggest future directions in both aspects of data collection and model development.

In terms of data collection, more realistic collection strategies should be adopted.
In the collection process of QuAC, for example, workers are encouraged to create \textit{long} conversations with \textit{few} unanswerable questions, causing the worker to ask questions conservatively and passively ask questions very likely answerable.
That may consequently result in the shorter distance between consecutive answers.
In addition, it is unknown that if such setting is realistic.
Is such information seeking pattern practical in real scenarios?
Therefore, inventing a more realistic data collection process may be an important future direction.

In terms of model development, conversational question answering systems should address more about semantic comprehension.
As we have shown, better performance does not necessarily imply better content comprehension, so proving the model's conversational understanding still remains challenging.
It is also unclear how to design a model that learns conversation comprehension naturally.
The future work could focus more studies on those directions, including generalization and robustness to diverse scenarios.

\section*{Acknowledgements}
We thank reviewers for their insightful comments. This work was financially supported from the Young Scholar Fellowship Program by Ministry of Science and Technology (MOST) in Taiwan, under Grant 108-2636-E002-003 and 108-2634-F-002-019.

\bibliographystyle{aaai}
\bibliography{aaai20.bib}

 \appendix

 \section{Examples of Attack on QuAC, Repeat Attack, and Shuffled CoQA}

 \begin{figure}[h!]
     \centering
     \includegraphics[width=\linewidth]{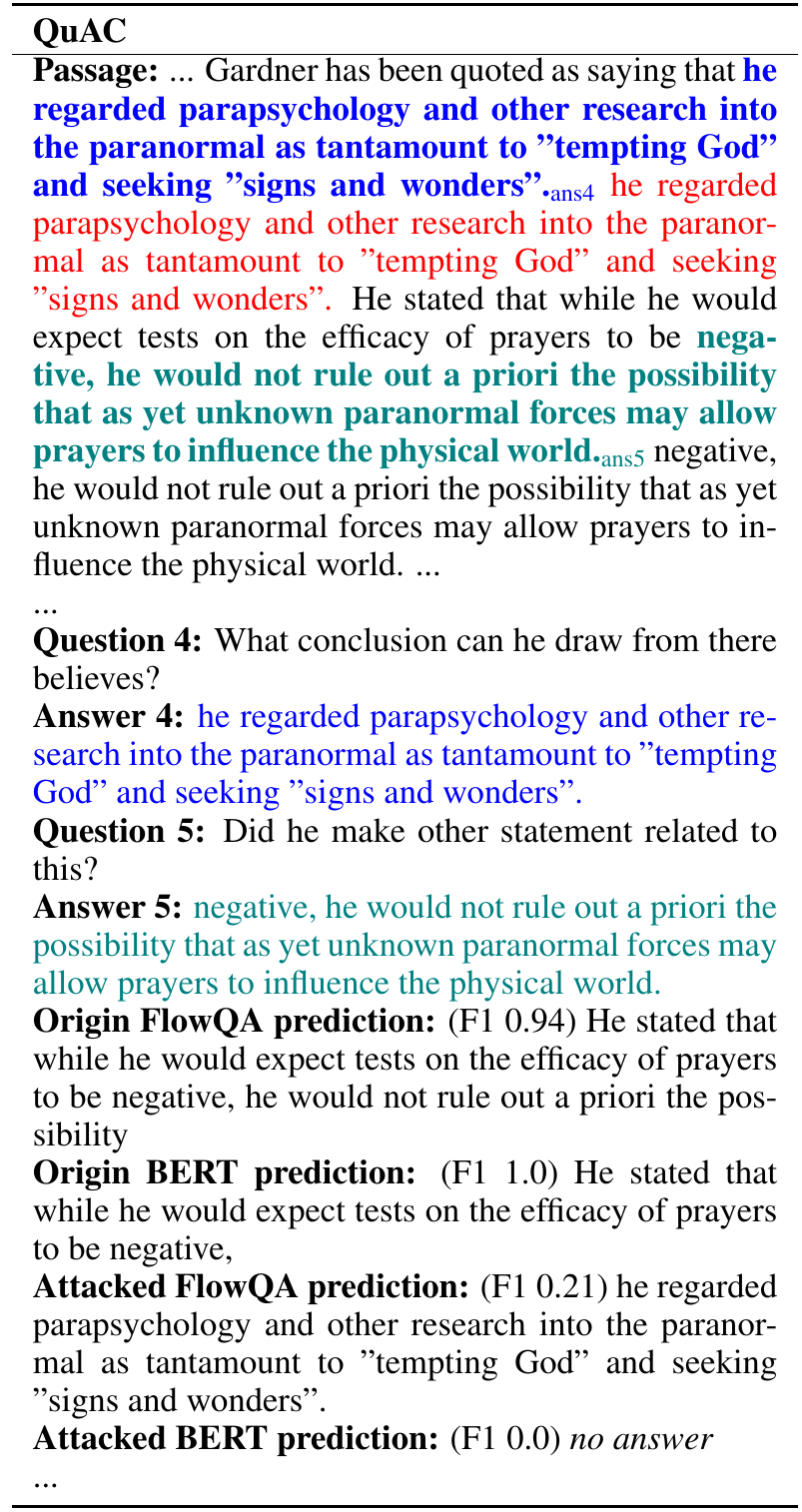}
     \caption{An example of repeat attack on QuAC and the model prediction which BERT predicts \textit{no answer}.}
     \label{tab:attack-quac-bert-na}
 \end{figure}

 \begin{figure}[t!]
     \centering
     \includegraphics[width=\linewidth]{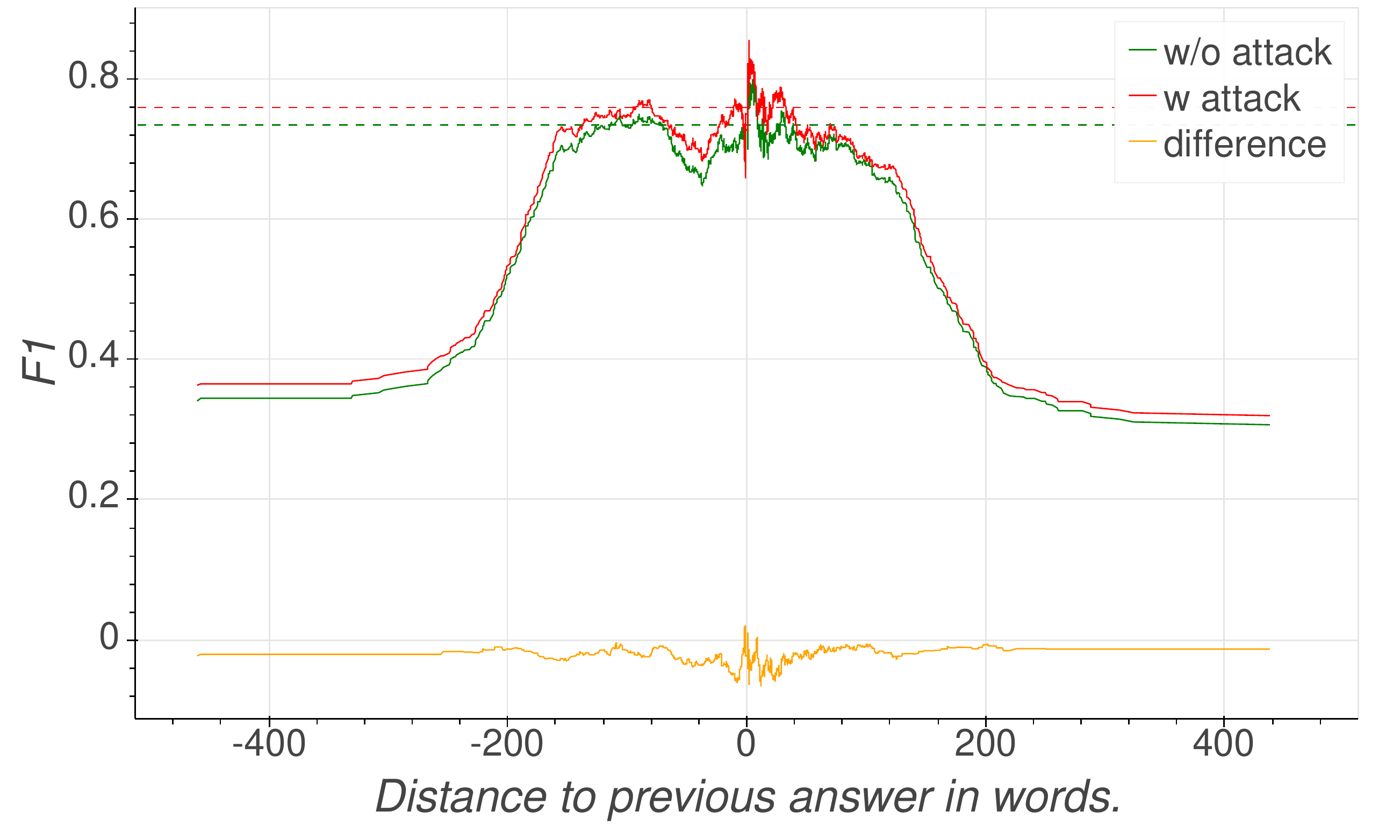}
     \includegraphics[width=\linewidth]{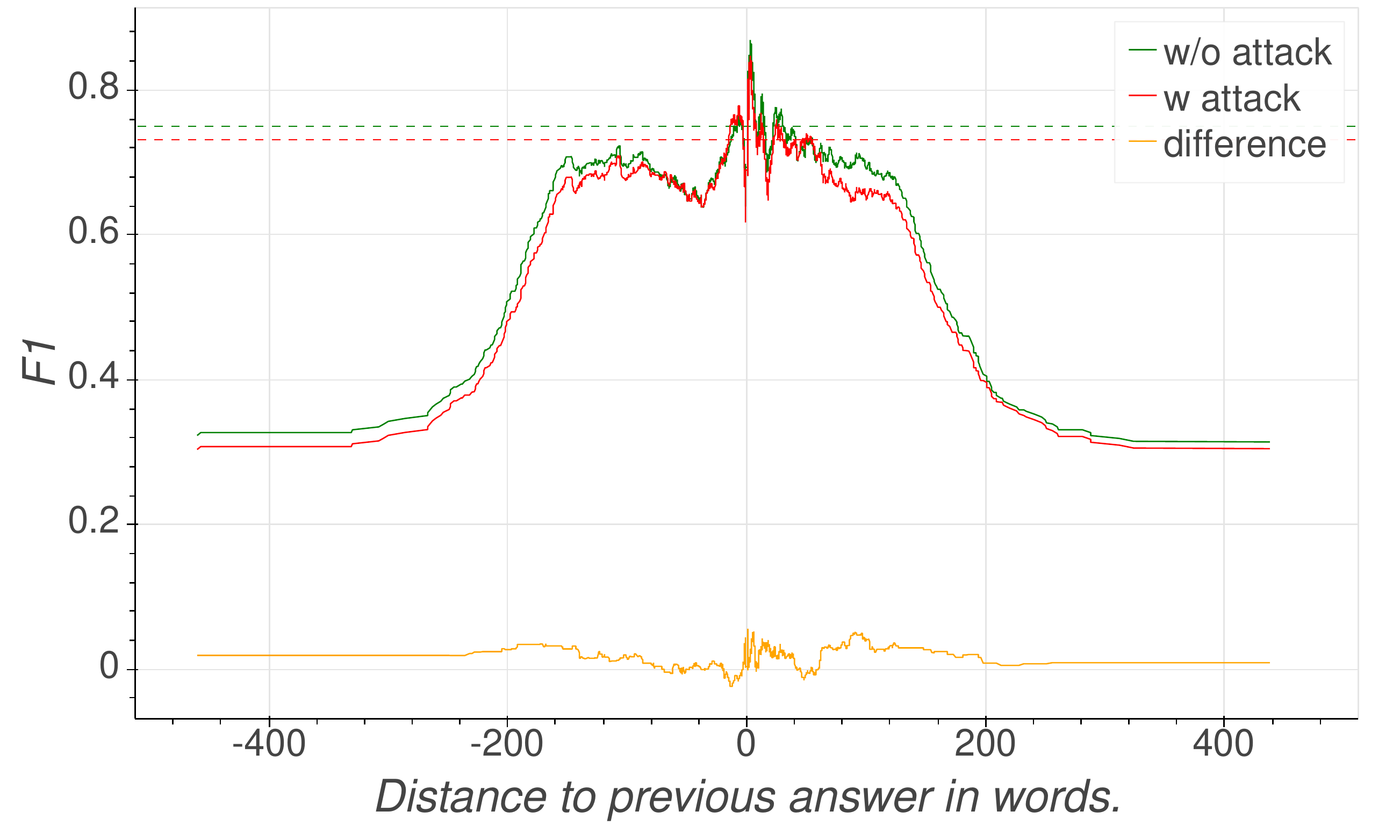}
     \caption{Relation between the F1 score before/after repeat attack and the answer’s distance to its previous answer on CoQA. The model used here is FlowQA.}
     \label{fig:coqa-attack-curve}
 \end{figure}


 \begin{figure}[t!]
     \centering
     \begin{tabular}{|p{0.9 \linewidth}|}
     \hline
     \textbf{Passage:} What are you doing, Cotton?!" The rest of her sisters were all orange with beautiful white tiger stripes like Cotton's mommy. Then Cotton thought, "I change my mind. All of her sisters were cute and fluffy, like Cotton. But Cotton wasn't alone in her little home above the barn, oh no. Cotton lived high up in a nice warm place above the barn where all of the farmer's horses slept. "I only wanted to be more like you". When her mommy and sisters found her they started laughing. "Don't ever do that again, Cotton!" So one day, when Cotton found a can of the old farmer's orange paint, she used it to paint herself like them. And with that, Cotton's mommy picked her up and dropped her into a big bucket of water. She shared her hay bed with her mommy and 5 other sisters. Her sisters licked her face until Cotton's fur was all all dry. Next time you might mess up that pretty white fur of yours and we wouldn't want that!" We would never want you to be any other way". Being different made Cotton quite sad. She often wished she looked like the rest of her family. When Cotton came out she was herself again. But she was the only white one in the bunch. Cotton's mommy rubbed her face on Cotton's and said "Oh Cotton, but your fur is so pretty and special, like you. they all cried. " I like being special". Once upon a time, in a barn near a farm house, there lived a little white kitten named Cotton. \\
     \hline
     \end{tabular}
     \caption{An example of random shuffled CoQA passage.}
     \label{fig:shuffled-coqa}
 \end{figure}

\end{document}